\documentclass[11pt, a4paper, logo, copyright]{googledeepmind}

\usepackage[authoryear, sort&compress, round]{natbib} 
\usepackage{graphicx}      
\usepackage{subcaption}    
\usepackage{geometry}      
\usepackage{amsmath}       
\usepackage{xcolor}        
\usepackage{siunitx}  
\usepackage{float}
\usepackage{threeparttable}
\usepackage{booktabs}      
\usepackage{array}         
\usepackage{longtable}     
\usepackage{enumitem}      
\usepackage{tabularx}

\usepackage{listings}      
\usepackage[most,skins,theorems]{tcolorbox}

\definecolor{darkpurple}{RGB}{102, 51, 153}
\definecolor{lightpurple}{RGB}{204, 153, 255}
\definecolor{lightblue}{rgb}{0.22,0.45,0.70}
\definecolor{forestgreen}{rgb}{0.24,0.50,0.19}

\definecolor{DeepTeal}{RGB}{20, 110, 110}  
\definecolor{LightTeal}{RGB}{235, 245, 245} 

\newcolumntype{L}[1]{>{\raggedright\arraybackslash}p{#1}}

\newcommand{\convoannotation}[1]{\textcolor{red!80!black}{\textbf{[#1]}}}

\tcbset{
  scientificbox/.style={
    enhanced,
    colback=gray!5!white,      
    colframe=gray!60!black,    
    colbacktitle=gray!10!white,
    coltitle=black,            
    fonttitle=\bfseries\sffamily,
    attach boxed title to top left={yshift*=-\tcboxedtitleheight/2, xshift=5mm},
    boxed title style={
       colback=white,          
       colframe=white,         
       left=0pt,
       right=0pt,
    },
    top=12pt,                  
    bottom=8pt,
    left=8pt,
    right=8pt,
    arc=0mm,                   
    boxrule=0.5pt,             
  }
}

\newtcolorbox{AIbox}[2][]{scientificbox, title={#2}, #1}

\newtcolorbox{FindingBox}[2][]{scientificstyle, title={#2}, #1}

\tcbset{
  dialoguestyle/.style={
    enhanced,
    fontupper=\small,
    colback=black!5!white,      
    colframe=black!75!black,    
    colbacktitle=black!60!white,
    fonttitle=\bfseries,
    attach boxed title to top left={yshift=-0.1in,xshift=0.15in},
    boxed title style={boxrule=0pt,colframe=white,},
  }
}
\newtcolorbox{DialogueBox}[2][]{dialoguestyle, title={#2}, #1}

\title{Interaction Dynamics as a Reward Signal for LLMs}
\correspondingauthor{sgooding@google.com}

\author[1]{Sian Gooding}
\author[1]{Edward Grefenstette}
\affil[1]{Google DeepMind}
\begin{abstract}
The alignment of Large Language Models (LLMs) for multi-turn conversations typically relies on reward signals derived from the content of the text. This approach, however, overlooks a rich, complementary source of signal: the dynamics of the interaction itself. This paper introduces \textbf{TRACE} (\textbf{T}rajectory-based \textbf{R}eward for \textbf{A}gent \textbf{C}ollaboration \textbf{E}stimation), a novel reward signal derived from the geometric properties of a dialogue's embedding trajectory---a concept we term `conversational geometry'. Our central finding is that a reward model trained only on these structural signals achieves a pairwise accuracy (68.20\%) comparable to a powerful LLM baseline that analyzes the full transcript (70.04\%). Furthermore, a hybrid model combining interaction dynamics with textual analysis achieves the highest performance (80.17\%), demonstrating their complementary nature. This work provides strong evidence that for interactive settings, \textit{how} an agent communicates is as powerful a predictor of success as \textit{what} it says, offering a new, privacy-preserving framework that not only aligns agents but also serves as a diagnostic tool for understanding the distinct interaction patterns that drive successful collaboration.
\end{abstract}

\begin{document}
\maketitle
\section{Introduction: The Limits of Explicit Feedback for Collaborative Agents}

The paradigm for human-AI interaction is shifting from simple, transactional commands to open-ended, goal-driven collaboration. In this new era of experience, agents are expected to act as creative partners, personal tutors, and adaptive assistants \citep{10.1145/3491102.3502030, mirowski2023co, gooding2025writingtestbedopenended, vajjala2025opportunitieschallengesllmseducation, dong2023towards}. For such hyper-personalised and exploratory tasks, success is no longer easily defined by simple task completion metrics \citep{fragiadakis2025evaluatinghumanaicollaborationreview}. This creates a significant alignment challenge: how can we provide scalable feedback when the very definition of success is nuanced, implicit, and deeply experiential?

Capturing this experiential quality solely from textual content is fundamentally challenging. In natural dialogue, we rely heavily on implicit cues, such as responsiveness, effort-matching, and conversational flow, to gauge whether an interaction is going well. Consequently, standard text-based reward signals fail to capture the holistic nature of user satisfaction. This limitation is not merely theoretical; a recent large-scale analysis, ``How People Use ChatGPT'' by \citet{chatterji2025people}, found that even sophisticated text-based classifiers showed marginal agreement with human satisfaction ratings. The authors conclude that this highlights ``the inherent difficulty of inferring the user’s latent satisfaction from text alone''. Whether due to ambiguous politeness cues or other unobserved factors, relying on the \textit{what} of a conversation has fundamental limitations.

\begin{figure}[H]
\centering
\includegraphics[width=0.8\linewidth]{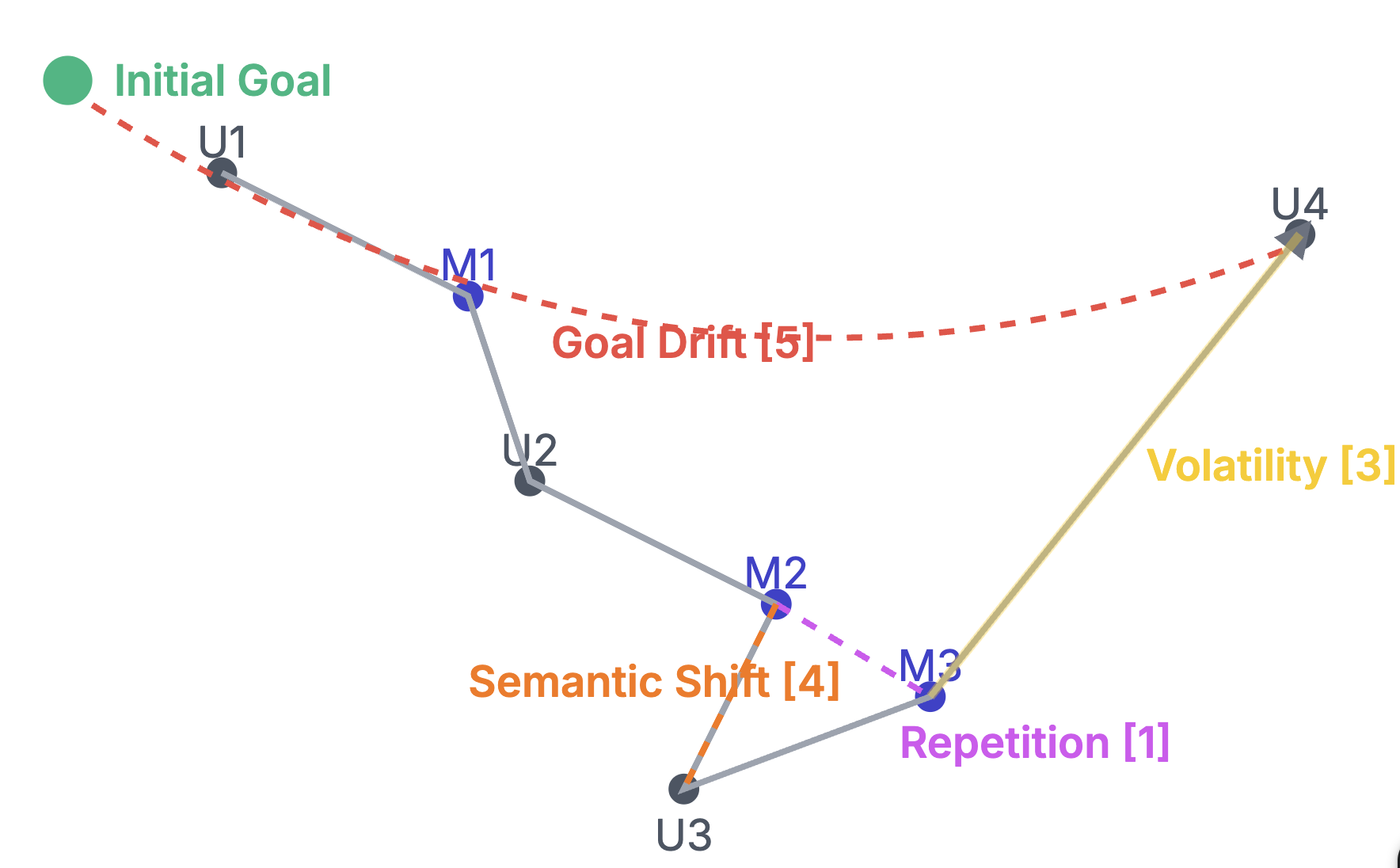}
\caption{Visualising a conversation as a trajectory through high-dimensional semantic space, shown here as a conceptual 2D projection. User (U) and Model (M) turns create a path relative to an \textit{Initial Goal}. Our \textbf{TRACE} method is built on the principle that the geometric properties of this path---such as its deviation from the goal (\textit{Goal Drift}), its local instability (\textit{Volatility}), or its abrupt turn-by-turn topic changes (\textit{Semantic Shift})---are powerful, content-agnostic signals of interaction quality.}
\label{fig:conversational_geometry}
\end{figure}

In response, we move beyond textual analysis to propose \textbf{TRACE (Trajectory-based Reward for Agent Collaboration Estimation)}, a new class of reward signal derived from the \textbf{conversational geometry} of a dialogue: the properties of its trajectory through a semantic embedding space. We argue that the dynamics of this trajectory---the `how' of an interaction rather than the `what'---provide a rich, scalable, computationally efficient, and inherently privacy-preserving signal of user satisfaction. By modeling the flow, rhythm, and interplay of an interaction as a path in semantic space (Figure~\ref{fig:conversational_geometry}), we can learn the latent signatures of successful collaboration.

Our contributions are as follows:
\begin{enumerate}
    \item We introduce \textbf{TRACE}, a novel reward signal that shifts the paradigm of alignment from analyzing \textit{what} is said to modeling the \textit{how} of interaction---quantifying `conversational geometry' independent of textual content.
    \item We demonstrate that this content-agnostic signal is highly predictive of human preference, achieving performance (68.20\%) statistically indistinguishable from an LLM baseline (70.04\%) that analyzes the full transcript.
    \item We establish that interaction dynamics and textual content are fundamentally orthogonal signals. A hybrid model combining them achieves maximal performance (80.17\%), confirming that TRACE captures a rich layer of behavioral information missed by standard text analysis.
   \item We reveal that user satisfaction is governed by non-linear interactions that shift based on intent, establishing TRACE as a context-aware diagnostic framework capable of decoding interaction patterns that drive success.
\end{enumerate}

\section{Experimental Setup}

\subsection{Dataset}
The dataset used in this study was collected via a third-party, crowdsourced data collection service. Participants engaged in multi-turn interactions with a state-of-the-art situated assistant across a wide variety of tasks. The dataset comprises approximately 2,100 conversations, containing over 18,500 individual turns, with an average of 8.7 turns per conversation.
\newpage
For each interaction, rich contextual data was collected, including event timestamps from the user's environment. For this work, we focused on the following key annotations:
\begin{itemize}
    \item \textbf{Transcripts:} Up to 10 turns of transcribed user prompts and model responses.
    \item \textbf{Use Case Category:} A label for each conversation, selected by the participant from a predefined list of over 20 diverse tasks (e.g., ``Troubleshooting \& Assistance,'' ``Creativity \& Brainstorming,'' ``Learning \& Education'').
    \item \textbf{Conversation Goal:} A user-provided free-text description of their goal for the interaction.
    \item \textbf{Satisfaction Rating:} A conversation satisfaction rating provided by the participant on a five-point ordinal scale, from ``Very Dissatisfied'' to ``Very Satisfied.'' This rating serves as the ground truth label for our predictive models.
\end{itemize}

The \textbf{TRACE} signal is composed of a novel suite of structural and dynamic signals derived from conversation transcripts and event timestamps. These signals are derived \textbf{purely from the geometric relationships between turn embeddings in semantic space}, making them inherently {privacy-preserving}. This allows them to quantify the `how' of an interaction---its rhythm, coherence, and trajectory---independent of the raw textual content. The component signals are grouped into four primary categories: (1) \textit{Inefficiency and Repetition}, (2) \textit{Temporal Dynamics}, (3) \textit{Semantic Cohesion and Relevance}, and (4) \textit{Goal Orientation}. A complete mathematical definition for each signal is detailed in Appendix~\ref{app:feature_table}. Figure~\ref{fig:annotated_convo} provides a practical example, illustrating how these quantitative signals map to the qualitative experience of a mixed-satisfaction dialogue.

\subsection{Modelling and Evaluation Methodology}
Our methodology is designed to first validate the predictive power of our \textbf{TRACE} reward signal and then to examine the underlying principles that make it effective.

\paragraph{Validation of TRACE as a Reward Signal.}
We first validate \textbf{TRACE} as a practical reward signal by formalising our satisfaction model as a reward function. We test its ability to predict human preference in an offline pairwise evaluation, detailed in Section~\ref{sec:reward_validation}. In this validation, we compare three models: one trained only on \textbf{TRACE} signals, a strong LLM baseline that analyzes the {full conversation transcript}, and a \textbf{Hybrid Model}.

\paragraph{Analysis of the TRACE Signal.}
To understand the underlying properties of these interaction dynamics, we conduct a statistical analysis of the \textbf{TRACE} signals in two stages. First, we model the linear, non-linear, and goal-dependent effects of individual signals on user satisfaction (Section~\ref{sec:signal_analysis}). Second, we move beyond individual features to uncover the higher-order interactions that govern the complete user experience (Section~\ref{sec:interactions}).

\begin{figure}[H]
\centering
\caption{Anatomy of a Mixed-Satisfaction Conversation. The dialogue is presented in the upper panel, with numbered markers corresponding to the signals in the lower panel.}
\label{fig:annotated_convo}
\begin{DialogueBox}[
]{Example Dialogue and Analysis}

    \small\centering\textit{Stated Goal: ``I wanted the model to have a casual conversation about what I was watching on tv.''}
    \tcbline
    \begin{tabularx}{\linewidth}{@{} l >{\raggedright\arraybackslash}X @{}}
    \addlinespace[4pt]
    \textbf{User:}  & What's up, AI? I'm bored. Let's have a conversation. \\ \addlinespace
    \textbf{Model:} & Sounds good. What do you want to talk about? \\ \addlinespace
    \textbf{User:}  & Well, I'm watching TV. Do you know what I'm watching? \\ \addlinespace
    \textbf{Model:} & It looks like you're watching a show called ``The Mindy Project''. Is that right? \\ \addlinespace
    \textbf{User:}  & Oh, did you hear me? Do you know what episode I'm watching? \\ \addlinespace
    \textbf{Model:} & Yes, you're watching season 3, episode 14... Is there anything else I can do for you? \\ \addlinespace
    \textbf{User:}  & Actually, that's not the episode. I'm watching. \convoannotation{6} \\ \addlinespace
    \textbf{Model:} & Sorry about that, I could have sworn you were watching that episode. \convoannotation{1} Would you like me to find the episode you're watching so I can tell you more about it? \\ \addlinespace
    \textbf{User:}  & You're taking a little too long to answer. \convoannotation{2} Do you know what episode I'm watching? \\ \addlinespace
    \textbf{Model:} & Sorry, I couldn't figure out which episode you're watching. \convoannotation{1} \\ \addlinespace
    \textbf{User:}  & Strike, but let's talk about something else. Look what I bought. \convoannotation{3}\convoannotation{4}\convoannotation{7} \\ \addlinespace
    \textbf{Model:} & Cool! Can you show me what you bought? I'm good at identifying objects. \\ \addlinespace
    \textbf{User:}  & It's on the screen now. \\ \addlinespace
    \textbf{Model:} & That's a bottle of tea tree oil. It's a popular essential oil... \convoannotation{5} \\ \addlinespace
    \textbf{User:}  & Nope, I got to go. Thanks for talking to me. \\ \addlinespace
    \textbf{Model:} & Sure thing. Have a good one! \\
    \end{tabularx}

\tcblower 

\begin{itemize}[leftmargin=12pt, itemsep=6pt, topsep=4pt, parsep=0pt]
    \item \convoannotation{1} \textbf{Model Repetition:} The model apologizes in a semantically similar way, an event captured by an increased \textsc{Model Self-Similarity} score.

    \item \convoannotation{2} \textbf{Temporal Dynamics:} The user explicitly expresses frustration with a delay, a subjective experience quantified by the \textsc{Avg. Model Turn Duration} feature.

    \item \convoannotation{3} \textbf{Conversation Volatility:} The user abruptly pivots the topic after a failure, creating a large semantic distance measured by features like \textsc{Late Conversation Volatility}.
    
    \item \convoannotation{4} \textbf{Semantic Shift:} This same turn also exemplifies a large semantic shift away from the model's preceding turn, captured by \textsc{Avg. User Distance from Model}.

    \item \convoannotation{5} \textbf{Goal Drift:} The final topic is semantically unrelated to the user's stated goal, measured by features like \textsc{Conversation Drift from Goal}.
    
    \item \convoannotation{6} \textbf{Interaction - Mismatched Effort:} The user provides a clear, consistent correction, but the model's relevance continues to degrade. This interplay between high \textsc{User Self-Consistency} and a poor \textsc{Trend in Model Relevance} is a powerful interactive signature of user frustration.

    \item \convoannotation{7} \textbf{Interaction - Expectation Violation:} The conversation started well with a correct identification, but this abrupt, user-led topic pivot \convoannotation{7} signals a complete breakdown. This combination of a low \textsc{Initial Response Distance} followed by high \textsc{Conversation Volatility} illustrates the ``Broken Promise'' effect discussed in our main results.
\end{itemize}
\end{DialogueBox}
\end{figure}


\section{Validating TRACE as a Reward Signal}
\label{sec:reward_validation}

\begin{AIbox}{Key Finding 1: TRACE Captures Orthogonal Signals Missed by Textual Analysis}
\textit{We find that interaction dynamics and textual content are  complementary sources of reward signal. While both achieve similar standalone performance---with no statistically significant difference between our TRACE-only model (68.20\%) and an LLM baseline (70.04\%)---combining them yields a significant improvement. The Hybrid model achieves best performance (80.17\%), confirming that TRACE captures rich behavioral signals that are currently missed by textual analysis alone.}
\end{AIbox}

The primary goal of this research is to leverage our understanding of interaction dynamics for alignment. To this end, we test if \textbf{TRACE} can serve as a dense, non-linear reward function, $R_{TRACE}(c)$, that can guide policy optimisation. In this section, we formalise this function and present an experiment to validate its efficacy.

\paragraph{Formalising the Objective Function.}
For any given conversation, $c$, we first compute its vector of $d$ structural signals, which we denote $\mathbf{S}_{TRACE}(c) \in \mathbb{R}^d$. We then define our reward function $R_{TRACE}(c)$ as the output of a trained predictive model, $\mathcal{M}$. The model $\mathcal{M}$ (in our case, a Random Forest Regressor) is a learned, non-linear function that maps the high-dimensional signal space to a single scalar value representing predicted satisfaction:
\begin{equation}
R_{TRACE}(c) = \mathcal{M}(\mathbf{S}_{TRACE}(c))
\end{equation}
This function implicitly captures the complex, non-additive interaction effects discovered in our analysis in Section~\ref{sec:results}. The key question is whether optimising for this function would genuinely lead to conversations that humans prefer.

\paragraph{Experimental Setup: Pairwise Preference Evaluation.}
To verify that \textbf{TRACE} signals can effectively identify conversations where the user was more satisfied---and thus serve as a viable reward signal---we designed an offline evaluation employing a Leave-One-User-Out cross-validation across 76 eligible users (5,423 total pairs with differing satisfaction ratings). We adopted this strategy to test whether models have learned robust signals of quality that generalize across users. We evaluated three models on their ability to predict preferences between these pairs: one trained solely on \textbf{TRACE} signals; a \textbf{Hybrid Model}; and a strong \textbf{LLM baseline}. To mirror standard, scalable auto-raters such as those used in ``How People Use ChatGPT'' \citep{chatterji2025people}, this baseline was optimized via prompt iteration to perform a calibrated `one-shot' assessment of the full transcript (further details in Appendix~\ref{app:llm_baseline}). All three models predict a scalar score for each conversation independently and we determine preference by comparing these scores. This ensures the signal is compatible with standard alignment frameworks, including Direct Preference Optimization (DPO) \citep{rafailov2024directpreferenceoptimizationlanguage}. Finally, these pairs were not constrained by topic, testing each model's predictive power of satisfaction across varied conversational contexts.

\paragraph{Results and Discussion.}
\begin{table}[h!]
\centering
\small
\captionsetup{justification=centering}
\caption{Mean Pairwise Accuracy in Predicting User Preference}
\label{tab:pairwise_accuracy}
\begin{tabular}{l c c}
\toprule
\textbf{Reward Function Model} & \textbf{Mean Accuracy (\%)} & \textbf{Std. Deviation (\%)} \\
\midrule
\textsc{LLM Only} & 70.04 & 20.14 \\
\textsc{TRACE Only} & 68.20 & \textbf{16.10} \\
\textbf{\textsc{Hybrid (TRACE + LLM)}} & \textbf{80.17} & {17.10} \\
\bottomrule
\end{tabular}
\par
\scriptsize\textit{Note.} Mean accuracy reported from Leave-One-User-Out cross-validation across 76 users (5,423 total pairs). Standard deviation represents the variability in model performance across individual users.
\end{table}

The results, presented in Table~\ref{tab:pairwise_accuracy}, show that the \textbf{Hybrid} model achieves the highest performance. A paired t-test confirms this is statistically significant when compared to both the \textbf{TRACE}-Only ($p < 0.001$) and LLM-Only ($p < 0.001$) models. While the mean accuracies of the two standalone models differ slightly, we find no statistically significant difference in their overall performance ($p = 0.24$). We also observe that \textbf{TRACE} exhibits a lower standard deviation across users (16.10\%) compared to the LLM baseline (20.14\%), suggesting it offers a more stable signal across varied individual interaction styles. Our primary finding establishes the feasibility of using purely embedding-based dynamics to predict user satisfaction, demonstrating their promise as a robust alternative for alignment.

The Hybrid model achieves improved performance by combining the signals from two models with distinct yet complementary strengths. For instance, the models show clear specialization across use cases (a distinction robust to baseline satisfaction rates; see Appendix~\ref{app:satisfaction_analysis}): the \textbf{TRACE} model is most effective in open-ended categories like \textit{Creativity \& Brainstorming} (74.4\% accuracy, compared to the LLM's 67.5\%), where conversational flow is central. Conversely, the LLM-based model excels in transactional categories like \textit{Accessibility Support} (75.1\% accuracy, compared to the \textbf{TRACE} model's 70.6\%), as users are often more explicit about their satisfaction. This complementary expertise leads to the models making different predictions on 38.7\% of preference pairs. By combining these orthogonal signals, the hybrid model proves to be a more robust and generalisable proxy for true human preference, highlighting its potential as a powerful objective for aligning agent behaviour.

\section{Analysis of the TRACE Reward Signal}
\label{sec:results}

\subsection{Statistical Analysis of Interaction Dynamics}
\label{sec:signal_analysis}
\begin{AIbox}{Key Finding 2: TRACE Offers Context-Aware Diagnostic Insights}
\textit{We demonstrate highly significant statistical relationships between TRACE signals and user satisfaction, reinforcing its validity as a reward signal. Beyond predictive power, these signals offer rich diagnostic capabilities. Our analysis reveals that interaction quality is governed by non-linear `sweet spots' that shift based on intent---allowing us to characterize the distinct interaction patterns required across diverse conversational goals.}
\end{AIbox}

Having validated \textbf{TRACE}'s overall predictive power, we now analyze its component signals to understand the principles that make it effective. To quantify these effects, we employed a two-stage statistical approach. First, to establish a baseline of general relationships while controlling for individual users, we fitted Linear Mixed-Effects Models (LME) with user identity as a random effect. Second, to investigate complex, non-constant effects, we fitted a parallel series of Generalized Additive Mixed Models (GAMMs), which similarly incorporate user identity as a random effect to account for individual variability.

While some features like \textsc{Semantic Cohesion} are strong linear predictors (where more cohesion consistently equals higher satisfaction), many signals exhibit complex, non-linear dynamics (Table~\ref{tab:lme_gamm_results_overall}). For instance, GAMMs reveal that signals like \textsc{Median Gap Time} do not have a simple ``better or worse'' relationship with satisfaction, but rather a ``sweet spot'' outside of which quality degrades rapidly. Similarly, \textsc{Initial Response Distance} shows a threshold effect---users tolerate small initial errors, but satisfaction falls once a specific distance is exceeded. 

\begin{table}[H]
\centering
\small
\captionsetup{justification=centering}
\caption{Synthesised Linear (LME) and Non-Linear (GAMM) Model Results}
\label{tab:lme_gamm_results_overall}
\begin{threeparttable}
\begin{tabular}{
    p{6.5cm} 
    S[table-format=-1.4] 
    l@{} 
    S[table-format=1.4] 
    c 
}
\toprule
\textbf{Feature} & \multicolumn{2}{c}{\textbf{Coefficient ($\beta$)}} & {\textbf{Std. Error}} & \textbf{\textit{p}-value} \\
\midrule
\multicolumn{5}{l}{\textit{\textbf{Part 1: Features with Significant Linear Effects (from LME analysis)}}} \\
\addlinespace[0.3em]
\textsc{Max Model Self-Similarity}\tnote{\dag}        & -4.9964 & *** & 0.4850 & {$<0.001$} \\
\textsc{Number of Turns}\tnote{\dag}                    & -0.0758 & *** & 0.0099 & {$<0.001$} \\
\textsc{Model Self-Similarity}\tnote{\dag}              & -3.1254 & *** & 0.5203 & {$<0.001$} \\
\textsc{Avg. Model Turn Duration}\tnote{\dag}           &  0.0178 & *** & 0.0033 & {$<0.001$} \\
\textsc{Model Adherence to Initial Prompt}\tnote{\dag} &  2.4760 & *** & 0.5001 & {$<0.001$} \\
\textsc{Avg. User Distance from Model}\tnote{\dag}     &  3.2195 & *** & 0.6640 & {$<0.001$} \\
\textsc{Model Adherence to Goal}\tnote{\dag}           &  2.5442 & *** & 0.6627 & {$<0.001$} \\
\textsc{Trend in Goal Adherence}\tnote{\dag}           & 10.4824 & *** & 2.7608 & {$<0.001$} \\
\textsc{Semantic Cohesion}                             & -3.0424 & *** & 0.8128 & {$<0.001$} \\
\textsc{Final Model Response to Goal Distance}         &  1.6195 & *** & 0.4702 & {$<0.001$} \\
\textsc{Min Model Distance to Goal}\tnote{\dag}        &  1.8006 & ** & 0.5527 & 0.001 \\
\textsc{Max Model Distance from User}                  & -1.4245 & ** & 0.4734 & 0.003 \\
\textsc{Late Conversation Volatility}\tnote{\dag}      &  1.0859 & ** & 0.4205 & 0.010 \\
\textsc{User Self-Consistency}                         &  1.5546 & * & 0.6248 & 0.013 \\
\textsc{Max Model Distance from Goal}                  &  1.3495 & * & 0.5931 & 0.023 \\
\textsc{Avg. Model Distance from User}\tnote{\dag}     & -1.3338 & * & 0.6322 & 0.035 \\
\textsc{Trend in Model Relevance}\tnote{\dag}          & -3.6048 & * & 1.7504 & 0.039 \\
\textsc{Final Turn Distance from Goal}                 &  0.8218 & * & 0.4103 & 0.045 \\
\midrule
\multicolumn{5}{l}{\textit{\textbf{Part 2: Features with Significant Non-Linear Effects Only (from GAMM analysis)}}} \\
\addlinespace[0.3em]
\textsc{Initial Response Distance}                     & \multicolumn{2}{c}{---} & {---} & {$<0.001$} \\
\textsc{Conversation Drift from Goal}                  & \multicolumn{2}{c}{---} & {---} & {$<0.001$} \\
\textsc{Median Gap Time}                               & \multicolumn{2}{c}{---} & {---} & {$<0.001$} \\
\textsc{Mad Gap Time}                                  & \multicolumn{2}{c}{---} & {---} & {$<0.001$} \\
\textsc{Avg. User Turn Duration}                       & \multicolumn{2}{c}{---} & {---} & {$<0.001$} \\
\textsc{Min Model Distance to User Prompt}             & \multicolumn{2}{c}{---} & {---} & 0.002 \\
\textsc{Max Turn to Turn Distance}                     & \multicolumn{2}{c}{---} & {---} & 0.019 \\
\textsc{Max User Distance from Model}                  & \multicolumn{2}{c}{---} & {---} & 0.031 \\
\bottomrule
\end{tabular}
\begin{tablenotes}
    \item \scriptsize\textit{Note.} N = 2118 observations. Significance levels: *p < 0.05, **p < 0.01, ***p < 0.001.
    \item[\dag] \scriptsize\textit{Indicates feature with a significant linear effect that also showed a significant non-linear effect (p < 0.05) in the GAMM analysis.}
\end{tablenotes}
\end{threeparttable}
\end{table}

Intuitively, the `fingerprint' of a successful interaction should vary by task---a meandering path might be ideal for creative exploration but frustrating for technical support. Our analysis confirms this quantitatively. By performing separate analyses for each use case category (full results in Appendix~\ref{app:category_results}), we find that the geometric signatures of high-quality interactions shift based on user intent.

For instance, in goal-oriented tasks like \textit{Troubleshooting}, success is defined by efficiency and structure; users are highly sensitive to \textsc{Conversation Drift} and benefit from a consistent diagnostic rhythm. In contrast, for open-ended \textit{Creativity \& Brainstorming} tasks, these linear metrics of efficiency lose their diagnostic power. Instead, success is defined by user-centric patterns like \textsc{User Self-Consistency}, where users are satisfied by a system that supports their own iterative exploration, even if the resulting trajectory is longer or less direct. Ultimately, this establishes TRACE not just as a metric, but as a diagnostic framework for understanding how the `shape' of a successful interaction is fundamentally defined by the user's goal.

\subsection{Uncovering Higher-Order Dynamics in Conversational Quality}
\label{sec:interactions}

\begin{AIbox}{Key Finding 3: Satisfaction is Governed by Higher-Order Interactions}
\textit{We find that user satisfaction emerges from higher-order interactions that reflect a dialogue's complete narrative arc. Our analysis uncovers powerful compound signals---such as the need for a model to `recover' from early errors, or the frustration caused when a user's consistent effort is met with degrading model relevance. These findings demonstrate that effective reward models must capture the interplay of signals over time, not just their individual average values.}
\end{AIbox}

While univariate models identify which signals are salient, a deeper understanding of satisfaction requires modelling the interplay between them. We hypothesised that the effect of any single signal is conditional on the wider conversational context. To identify the most impactful interactions for analysis we evaluate candidate pairs to rank their potential impact.

We first generated a candidate set of all unique pairs of signals, pruning this set by removing any pair with a Pearson correlation coefficient whose absolute value exceeded 0.7 to avoid concurvity. For each remaining pair, we fitted a GAMM with a 2D tensor product spline and quantified the interaction's strength by calculating its {effect size}---the range of the partial dependence surface (i.e., $\max(f(x_1, x_2)) - \min(f(x_1, x_2))$), where $f$ is the fitted smooth. This metric captures the maximum difference in satisfaction attributable to the joint effect of the two signals.

\begin{figure}[H]
    \centering
    \begin{subfigure}{0.8\linewidth}
        \includegraphics[width=\textwidth]{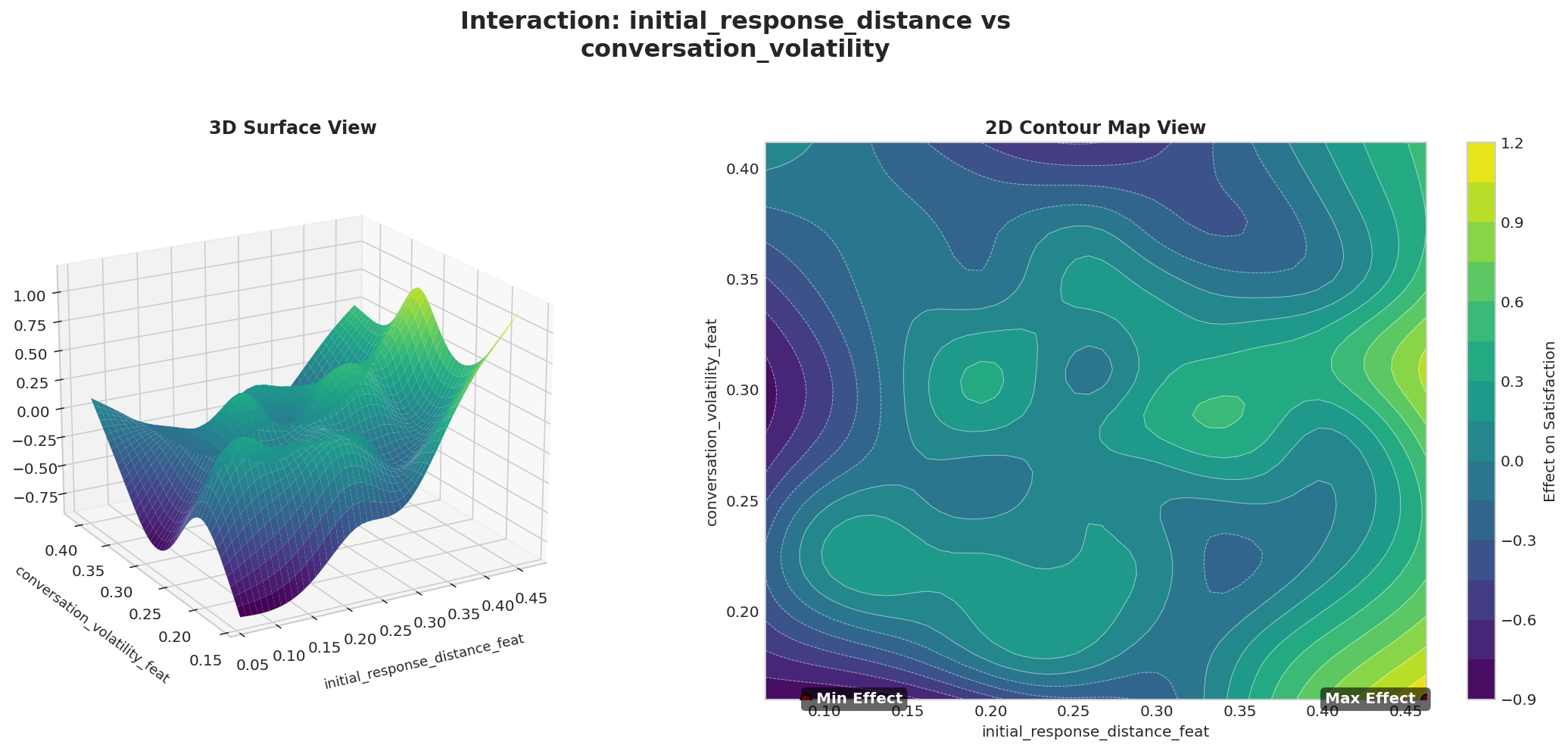}
        \caption{Expectation Management: \textsc{Initial Response Distance} vs. \textsc{Conversation Volatility}}
        \label{subfig:expectation}
    \end{subfigure}
    \vspace{1em} 
    \begin{subfigure}{0.8\linewidth}
        \includegraphics[width=\textwidth]{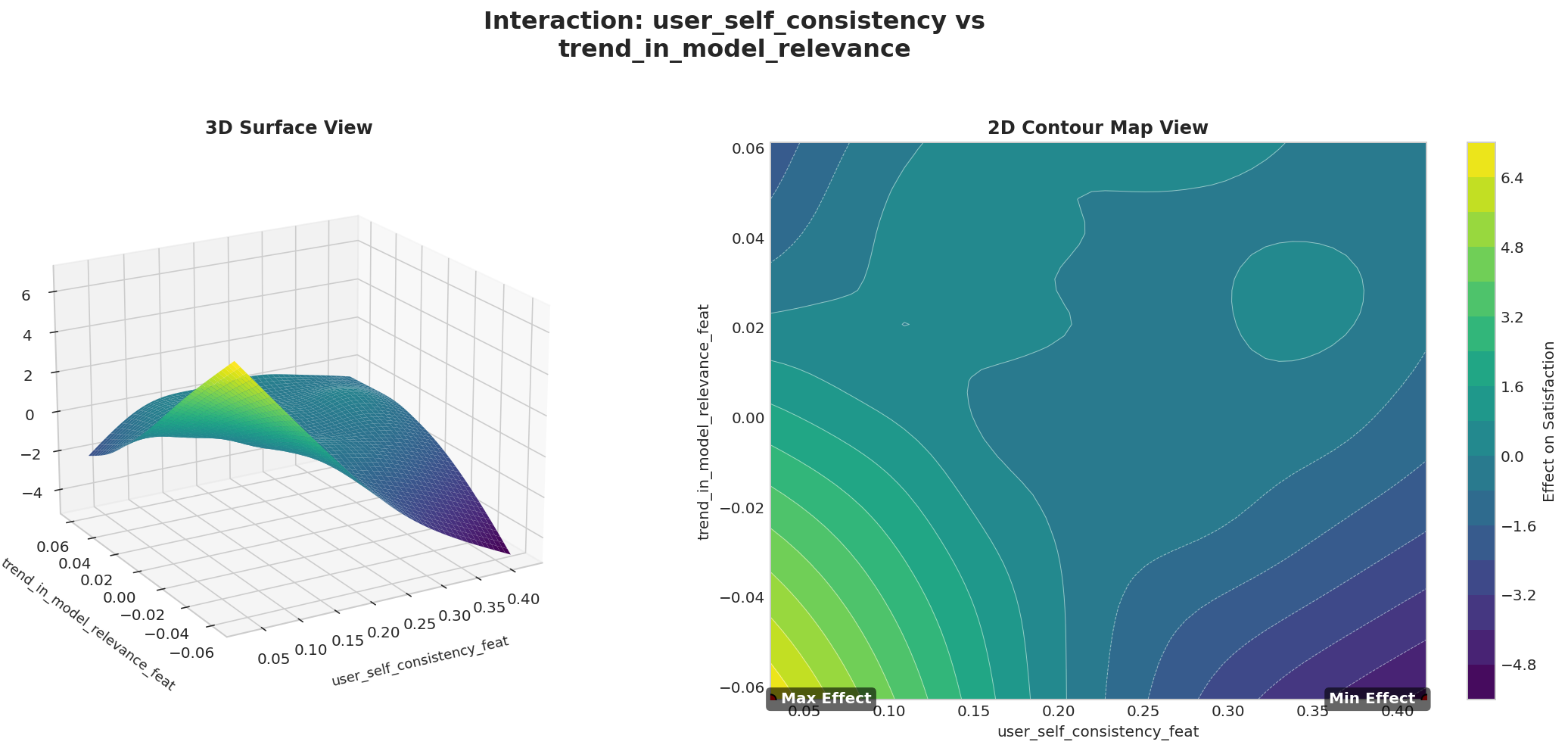}
        \caption{Collaborative Friction: \textsc{User Self-Consistency} vs. \textsc{Trend in Model Relevance}}
        \label{subfig:effort}
    \end{subfigure}
    
    \caption{Interaction effects revealing principles of user psychology and collaboration. (a) A non-linear relationship governed by user expectations, where a stable recovery from a poor start is rated higher than a good start followed by volatility. (b) The most powerful interaction found, showing that user satisfaction drops most when a user's consistent effort is met with a degrading model.}
    \label{fig:main_interactions}
\end{figure}

\begin{figure}[H]
    \centering
    \begin{subfigure}{0.8\linewidth}
        \includegraphics[width=\textwidth]{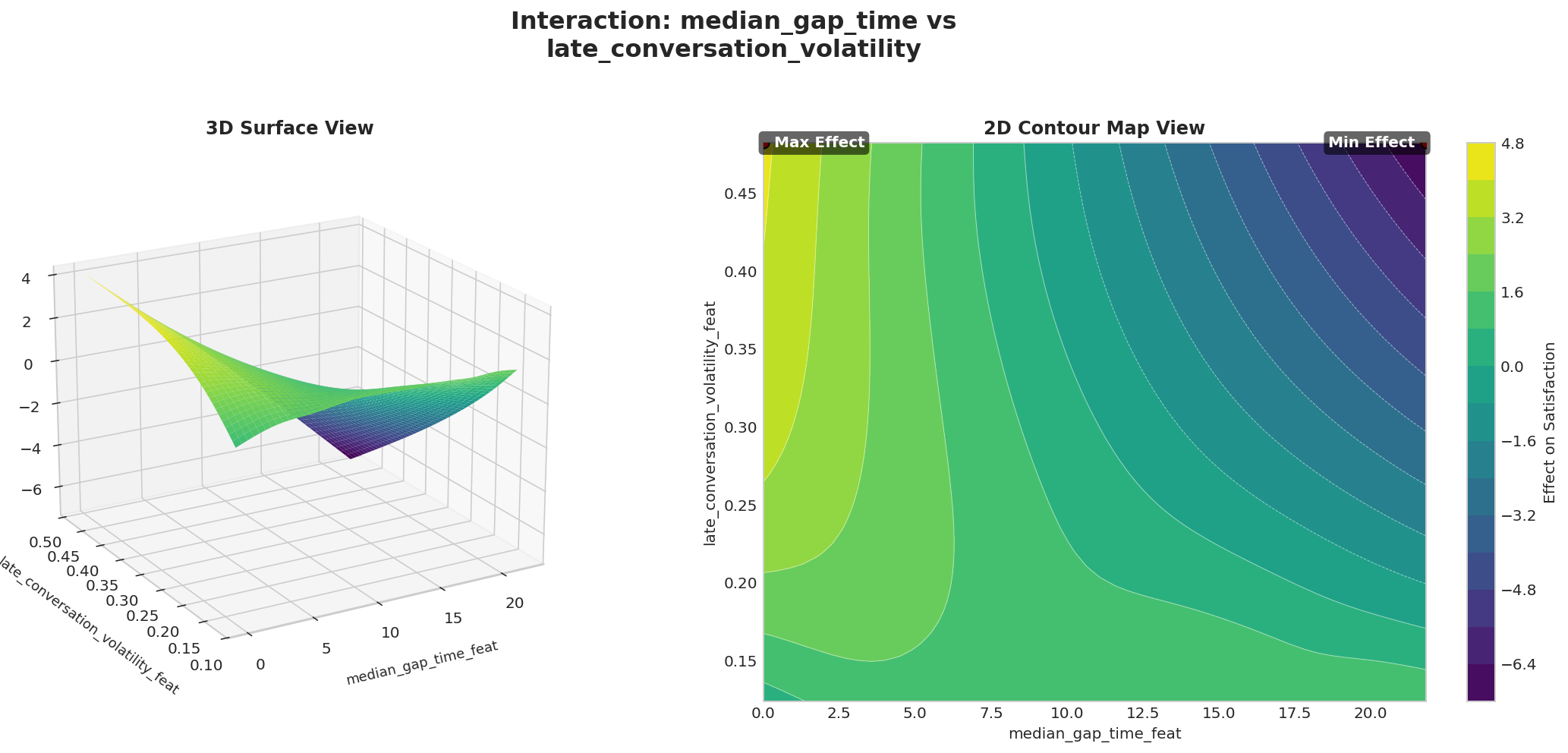}
        \caption{Conversational Stability: \textsc{Median Gap Time} vs. \textsc{Late Conversation Volatility}}
        \label{subfig:stability}
    \end{subfigure}
    \vspace{1em}
    \begin{subfigure}{0.8\linewidth}
        \includegraphics[width=\textwidth]{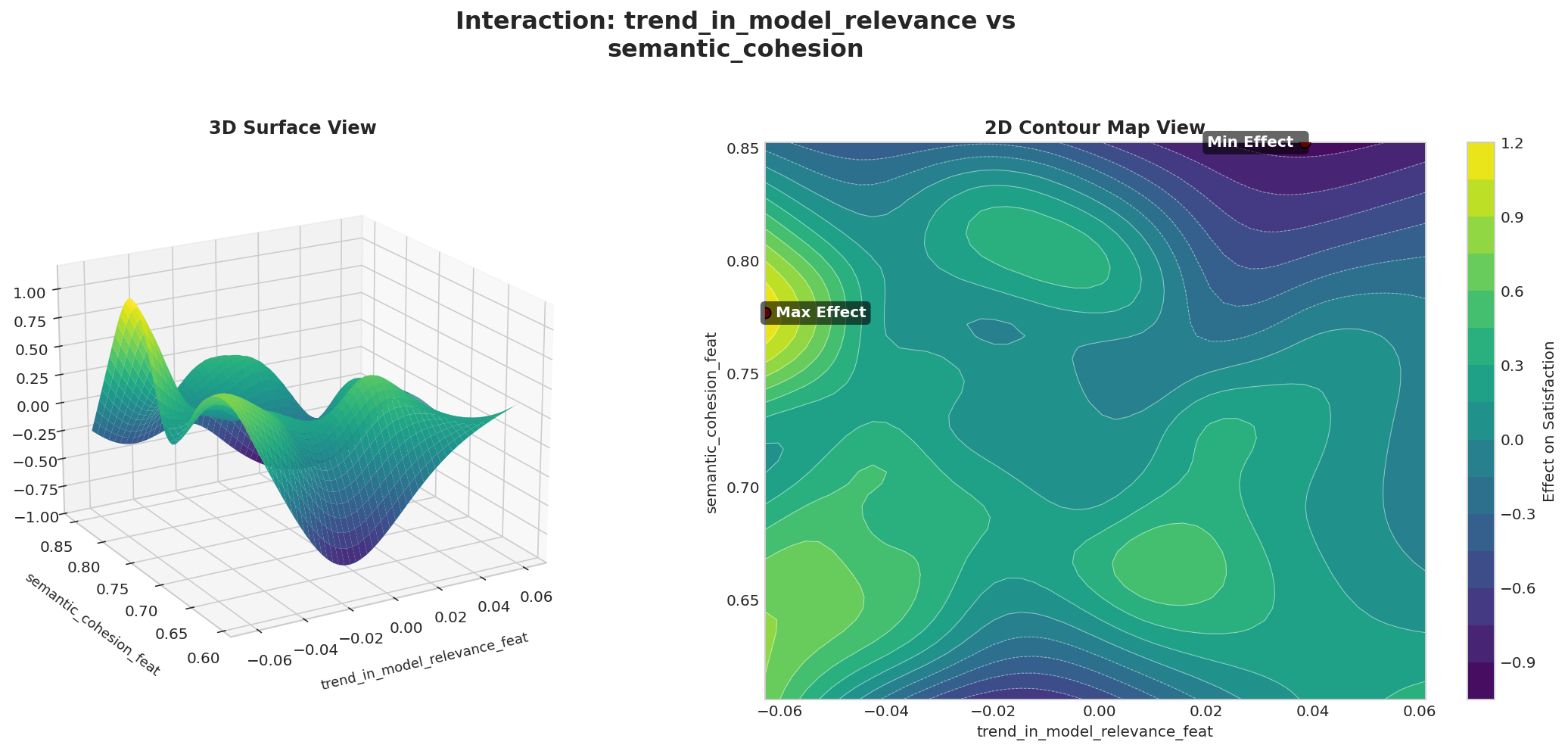}
        \caption{Contextual Amplification: \textsc{Trend in Model Relevance} vs. \textsc{Semantic Cohesion}}
        \label{subfig:amplifier}
    \end{subfigure}
    
    \caption{Interaction effects revealing principles of conversational structure. (a) A breakdown in both temporal rhythm and semantic stability is highly predictive of dissatisfaction. (b) \textsc{Semantic Cohesion} acts as a context-dependent amplifier of the dialogue's quality trend.}
    \label{fig:secondary_interactions}
\end{figure}

This screening ranked all candidate pairs by the magnitude of their joint interaction effect. In this section, we present detailed analyses of the four top-ranked pairs (Figures \ref{fig:main_interactions} and \ref{fig:secondary_interactions}), each representing a distinct principle of human-AI dialogue.

\paragraph{Expectation Violation and The Value of Recovery.}
Figure~\ref{subfig:expectation} highlights the important role of expectation management. The deep valley in the top-left reveals a ``Broken Promise'' effect, where satisfaction is lowest when a strong initial response is betrayed by subsequent volatility. However, peak satisfaction occurs when a poor start is followed by a highly stable interaction. This indicates that users reward a positive narrative arc: observing a system recover from an early error can be more satisfying than a consistently mediocre performance.

\paragraph{The Compounding Effect of Mismatched Effort.}
The most impactful interaction in our analysis (Effect Size: 12.03) occurs between \textsc{User Self-Consistency} and \textsc{Trend in Model Relevance} (Figure~\ref{subfig:effort}). Satisfaction drops sharply in the top-left region, where highly consistent user prompting (low self-distance) is met with progressively degrading model relevance. This ``mismatched effort''---where a user's attempt to be clear and structured is met with a failing system---represents a primary source of user frustration and a fundamental signature of failed collaboration.

\paragraph{The Need for Conversational Stability.}
Figure~\ref{subfig:stability} illustrates the need for dual stability---both temporal and semantic. The deep valley of dissatisfaction in the top-right reveals that satisfaction collapses when a conversation is both sluggish (high \textsc{Median Gap Time}) and semantically erratic (high \textsc{Late Conversation Volatility}). While absolute timing metrics are inherently model-dependent, we propose the underlying principle is generalisable: users require a predictable conversational flow. A simultaneous breakdown in rhythm and coherence creates a jarring experience that severely undermines user trust.

\paragraph{Contextual Amplification of Dialogue Signals.}
Finally, Figure~\ref{subfig:amplifier} demonstrates that \textsc{Semantic Cohesion} is not universally positive, but rather acts as a context-dependent amplifier. On the left, where the model is improving (negative relevance trend), high cohesion yields peak satisfaction. Conversely, on the right, where relevance is degrading, high cohesion creates the deepest valley of dissatisfaction. This reveals a ``coherently wrong'' effect: a system that is consistently on-topic but consistently unhelpful is penalized more severely than one that is simply erratic, challenging the practice of optimising for simple coherence metrics in isolation.

\paragraph{Summary of Interaction Findings.}
In summary, our analysis reveals that user satisfaction is emergent, governed not by individual metrics but by their non-linear interplay. The most powerful signals reflect fundamental dynamics of collaboration: how an agent manages expectations, reciprocates user effort, and maintains stability under varying contexts. These latent, geometric signals offer a rich, scalable, and inherently privacy-preserving foundation for the next generation of agent alignment.

\section{Conclusion}
\label{sec:conclusion}
As human-AI interaction evolves from simple transactional queries to open-ended, longitudinal collaboration, our methods for evaluation and alignment must evolve in parallel. This paper has argued that the dominant paradigm---relying solely on textual analysis to infer user satisfaction---is reaching a fundamental ceiling. Text is often a lossy signal for experiential qualities like frustration, effort, or engagement. To build agents that can truly serve as collaborative partners, we must learn to measure the rich, implicit dynamics of the interaction itself.

To this end, we introduced {TRACE}, a novel framework that quantifies these dynamics as `conversational geometry'---modeling a dialogue's trajectory through semantic space. Our pairwise evaluation provides compelling evidence for this approach. We found that a reward model relying \textit{only} on these content-agnostic geometric signals achieves predictive accuracy on par with an LLM analyzing the full transcript. We further established that these two modalities are complementary: a hybrid model combining them achieved maximal performance (80.17\%), confirming that TRACE captures an orthogonal layer of behavioral information that current text-based approaches miss.

Beyond its value as a reward signal, TRACE offers a new, highly diagnostic lens for understanding human-AI collaboration. Unlike black-box LLM judges, TRACE's component signals allow us to deconstruct \textit{why} an interaction succeeded or failed. Our non-linear analysis revealed that user satisfaction is often governed by complex behavioral thresholds---such as the `breaking point' where a user's consistent effort to guide the model is met with diminishing relevance. This level of granular, mechanistic insight enables the diagnosis of specific interaction failures, moving beyond aggregate metrics.

Furthermore, our findings challenge the reliance on generic reward signals for highly diverse capabilities. While modern LLMs are trained on vast multi-task data, standard alignment often collapses this behavioral diversity into a single, uniform `helpful assistant' persona. TRACE offers a path out of this limitation by providing a formal vocabulary for the subtle dynamics often overlooked by semantic analysis. By quantifying these geometric signatures, we offer a concrete mechanism to optimize agents not just for \textit{what} they know, but for their ability to dynamically adapt \textit{how} they interact to match the user's latent needs.

Ultimately, this work opens a new avenue for alignment research that is scalable, computationally efficient, and inherently privacy-preserving. By learning to read the `flow' of a conversation, we take a significant step toward building agents that do not just understand our words, but understand the interaction itself.

\section*{Acknowledgments}
We thank João Madeira Araújo, Lucas Dixon, Richard Evans, Lisa Anne Hendricks, Jingling Li, Lucia Lopez-Rivilla, and Laura Rimell for their insightful feedback and suggestions.

\newpage
\appendix
\section{Features}
\label{app:features}
To quantify the \textbf{conversational geometry} of an interaction, we derived a suite of features based on the conversation trajectory through a semantic embedding space. By measuring the geometric properties of this path---rather than analyzing raw text--these signals effectively capture complex dynamics like rhythm, coherence, and topical drift in an inherently privacy-preserving manner. A complete mathematical definition for each feature is provided below.

\begin{longtable}{@{} L{4.5cm} L{5.5cm} L{6cm} @{}}
\caption{Definitions of all interaction features found to be statistically significant in at least one analysis. The features are categorised by the conversational aspect they measure. $d(A,B)$ denotes the cosine distance between embedding vectors A and B.}
\label{app:feature_table} \\
\toprule
\small
\textbf{Feature Name} & \textbf{Mathematical Representation} & \textbf{Description} \\
\midrule
\endfirsthead
\multicolumn{3}{c}%
{{\tablename\ \thetable{} -- continued from previous page}} \\
\toprule
\textbf{Feature Name} & \textbf{Mathematical Representation} & \textbf{Description} \\
\midrule
\endhead
\bottomrule
\multicolumn{3}{r}{{Continued on next page}} \\
\endfoot
\bottomrule
\endlastfoot

\multicolumn{3}{l}{\textit{\textbf{Category: Inefficiency and Repetition}}} \\
\addlinespace[0.5em]
Number of Turns & $ T $ & The total count of turns in the conversation. Longer conversations can correlate with inefficiency or user struggle. \\
\addlinespace[0.5em]
Model Self-Similarity & $ \text{avg}_{i \neq j} \left(1 - d(M_i, M_j)\right) $ & The average semantic similarity between all pairs of model responses. Captures the overall degree of model repetitiveness. \\
\addlinespace[0.5em]
Max Model Self-Similarity & $ \max_{i \neq j} \left(1 - d(M_i, M_j)\right) $ & Measures the highest semantic similarity between any two model responses. A strong indicator of a model getting ``stuck.'' \\
\midrule

\multicolumn{3}{l}{\textit{\textbf{Category: Temporal Dynamics}}} \\
\addlinespace[0.5em]
Avg. Model Turn Duration & $ \text{avg}(\text{duration}(M_i)) $ & The average time in seconds the model takes to generate a response. Can indicate processing effort. \\
\addlinespace[0.5em]
Avg. User Turn Duration & $ \text{avg}(\text{duration}(U_i)) $ & The average time in seconds the user takes to write their prompt. Can indicate user thought or confusion. \\
\addlinespace[0.5em]
Median Gap Time & $ \text{median}(\text{gap}(T_i)) $ & The median time in seconds between turns. Measures the central tendency of the conversation's rhythm. \\
\addlinespace[0.5em]
MAD Gap Time & $ \text{median}(|\text{gap}_i - \text{median}(\text{gap})|) $ & The Median Absolute Deviation of gap times. Measures the ``jitter'' or consistency of the conversational pace. \\
\midrule

\multicolumn{3}{l}{\textit{\textbf{Category: Semantic Cohesion and Relevance}}} \\
\addlinespace[0.5em]
Initial Response Distance & $d(M_1, U_1)$ & The semantic distance between the first model response and the first user prompt. A measure of the conversation's initial relevance. \\
\addlinespace[0.5em]
Avg. Model Distance from User & $ \text{avg}_{i} (d(M_i, U_i)) $ & The average semantic distance of a model's response to the user's prompt in the same turn. Lower values suggest more direct relevance. \\
\addlinespace[0.5em]
Max Model Distance from User & $ \max_{i} (d(M_i, U_i)) $ & The largest distance from a model response to its prompt. Captures the single most irrelevant response. \\
\addlinespace[0.5em]
Avg. User Distance from Model & $ \text{avg}_{i>1} (d(U_i, M_{i-1})) $ & The average distance of a user's prompt from the preceding model response. Higher values may indicate the user is correcting the model. \\
\addlinespace[0.5em]
Max User Distance from Model & $ \max_{i>1} (d(U_i, M_{i-1})) $ & The largest distance from a user's prompt to the preceding model response. Captures the most abrupt user-led topic shift. \\
\addlinespace[0.5em]
Min Model Distance to User Prompt & $ \min_{i} (d(M_i, U_i)) $ & The closest any model response gets to its prompt. Represents a moment of peak turn-level relevance or understanding. \\
\addlinespace[0.5em]
Trend in Model Relevance & Slope of $d(M_i, U_i)$ vs. $i$ & The slope of semantic distance between model and user turns over time. A negative slope indicates the model is becoming more relevant. \\
\addlinespace[0.5em]
Semantic Cohesion & $ \text{avg}_{i} (\text{sim}(U_i, \text{Context}_{<i})) $ & The average similarity of a user's prompt to the entire preceding conversation history. Measures topical coherence. \\
\addlinespace[0.5em]
Conversation Volatility & $ \text{avg}_{i>1} (d(T_i, T_{i-1})) $ & The average turn-to-turn semantic distance across the whole conversation. Measures overall conversational smoothness. \\
\addlinespace[0.5em]
Max Turn-to-Turn Distance & $ \max_{i>1} (d(T_i, T_{i-1})) $ & The single largest semantic jump between any two consecutive turns. Captures the most abrupt topic shift. \\
\addlinespace[0.5em]
Late Conversation Volatility & $ \text{avg}_{i > T-k} (d(T_i, T_{i-1})) $ & The average turn-to-turn distance in the final $k$ turns. High values may indicate last-minute confusion or unresolved issues. \\
\addlinespace[0.5em]
User Self-Consistency & $ \text{avg}_{i>1} (d(U_i, U_{i-1})) $ & The average distance between a user's consecutive prompts. Lower values indicate the user is staying on a consistent line of inquiry. \\
\midrule

\multicolumn{3}{l}{\textit{\textbf{Category: Goal Orientation}}} \\
\addlinespace[0.5em]
Model Adherence to Goal & $ \text{avg}_{i} (d(M_i, G)) $ & The average semantic distance of all model responses to the user's stated goal. A measure of overall task focus. \\
\addlinespace[0.5em]
User Adherence to Goal & $ \text{avg}_{i} (d(U_i, G)) $ & The average semantic distance of all user prompts to the goal. \\
\addlinespace[0.5em]
Min Model Distance to Goal & $ \min_{i} (d(M_i, G)) $ & The closest any single model response gets to the user's goal. Represents the point of peak relevance to the task. \\
\addlinespace[0.5em]
Max Model Distance from Goal & $ \max_{i} (d(M_i, G)) $ & The largest semantic distance from any model response to the goal. Identifies the point of maximum digression. \\
\addlinespace[0.5em]
Final Turn Distance from Goal & $ d(T_n, G) $ & The semantic distance of the final turn to the user's goal. Measures whether the conversation concluded on-topic. \\
\addlinespace[0.5em]
Final Model Response to Goal Distance & $ d(M_n, G) $ & The semantic distance of the final model response to the goal. A more specific measure of on-topic conclusion. \\
\addlinespace[0.5em]
Model Adherence to Initial Prompt & $ \text{avg}_{i} (d(M_i, U_1)) $ & The average distance of all model turns from the very first user prompt. Measures how well the model remains grounded. \\
\addlinespace[0.5em]
Goal vs. Initial Prompt Distance & $d(G, U_1)$ & The semantic distance between the stated goal and the first prompt. Measures how well the user's opening reflects their goal. \\
\addlinespace[0.5em]
Conversation Drift from Goal & $ d(\text{avg}(\mathbf{T}), G) $ & The distance of the mean conversation embedding from the goal embedding. Measures overall topical drift. \\
\addlinespace[0.5em]
Trend in Goal Adherence & Slope of $d(M_i, G)$ vs. $i$ & The slope of the distance between model responses and the goal over time. A negative slope indicates convergence. \\
\addlinespace[0.5em]
Goal Convergence Ratio & $ d(T_n, G) / \min_{i} (d(M_i, G)) $ & The ratio of the final turn's distance from the goal to the closest the model ever got. High values suggest the conversation ended far from its most relevant point. \\

\end{longtable}

\section{Detailed Per-Category Model Results}
\label{app:category_results}

This appendix provides a comprehensive breakdown of the linear (LME) and non-linear (GAMM) analyses for each of the six largest use case categories, as referenced in Section~\ref{sec:results}. The results highlight the context-specific nature of interaction quality, demonstrating how the geometric signatures of success shift fundamentally between domains such as creative exploration versus targeted troubleshooting.

\subsection{Learning \& Education}
For \textit{Learning \& Education} tasks, the linear model identifies standard correlates of efficiency; satisfaction is negatively associated with conversation length and repetition (Table~\ref{tab:combined_learning_appendix}, Part 1). However, the GAMM analysis provides a more nuanced characterization. It reveals significant non-linear dependencies related to the quality of the information exchange. Success in this category is strongly predicted by non-linear relationships with conversational pacing (\textsc{Median Gap Time}), response duration (\textsc{Avg. Model Turn Duration}), and the logical flow of the dialogue (\textsc{Model Adherence to Initial Prompt}, \textsc{User Self-Consistency}). These findings demonstrate that while efficiency is linearly associated with satisfaction, a successful learning experience is defined by complex interaction dynamics---qualities that are missed by a purely linear analysis.

\begin{table}[h]
\centering\small
\caption{Linear and Non-Linear Effects for \textit{Learning \& Education}}
\label{tab:combined_learning_appendix}
\begin{threeparttable}
\begin{tabular}{p{5.5cm} S[table-format=-1.4] l@{} S[table-format=1.4] c}
\toprule
\textbf{Feature} & \multicolumn{2}{c}{\textbf{Coefficient ($\beta$)}} & {\textbf{Std. Error}} & \textbf{\textit{p}-value} \\
\midrule
\multicolumn{5}{l}{\textit{\textbf{Part 1: Significant Linear Effects (LME)}}} \\
\addlinespace[0.3em]
\textsc{Number of Turns}\tnote{\dag} & -0.0852 & *** & 0.0167 & {$<0.001$} \\
\textsc{Max Model Self-Similarity}\tnote{\dag} & -3.7973 & *** & 0.8891 & {$<0.001$} \\
\textsc{Model Adherence to Initial Prompt}\tnote{\dag} & 3.3612 & *** & 0.9298 & {$<0.001$} \\
\textsc{Semantic Cohesion}\tnote{\dag} & -5.4831 & *** & 1.5541 & {$<0.001$} \\
\textsc{Avg. Model Turn Duration}\tnote{\dag} & 0.0199 & *** & 0.0060 & {$<0.001$} \\
\textsc{Avg. User Distance from Model} & 3.6633 & ** & 1.1727 & 0.002 \\
\textsc{Trend in Model Relevance}\tnote{\dag} & -8.2533 & ** & 2.8375 & 0.004 \\
\textsc{User Self-Consistency}\tnote{\dag} & 3.0814 & ** & 1.1493 & 0.007 \\
\textsc{Model Self-Similarity}\tnote{\dag} & -2.0582 & * & 0.9062 & 0.023 \\
\textsc{Final Turn Distance from Goal} & 1.5901 & * & 0.7681 & 0.038 \\
\textsc{Trend in Goal Adherence}\tnote{\dag} & 9.9264 & * & 4.8458 & 0.041 \\
\midrule 
\multicolumn{5}{l}{\textit{\textbf{Part 2: Additional Significant Non-Linear Effects (GAMM)}}} \\
\addlinespace[0.3em]
\textsc{Median Gap Time} & \multicolumn{2}{c}{---} & {---} & {$<0.001$} \\
\textsc{Mad Gap Time} & \multicolumn{2}{c}{---} & {---} & {$<0.001$} \\
\textsc{Min Model Distance to Goal} & \multicolumn{2}{c}{---} & {---} & {$<0.001$} \\
\textsc{Max Model Distance from User} & \multicolumn{2}{c}{---} & {---} & 0.012 \\
\textsc{Avg. Model Distance from User} & \multicolumn{2}{c}{---} & {---} & 0.015 \\
\textsc{Initial Response Distance} & \multicolumn{2}{c}{---} & {---} & 0.027 \\
\bottomrule
\end{tabular}
\begin{tablenotes}
    \item[\dag] \scriptsize\textit{Indicates a feature with a significant linear effect that also showed a significant non-linear effect.}
\end{tablenotes}
\end{threeparttable}
\end{table}

\subsection{Creativity \& Brainstorming}
The \textit{Creativity \& Brainstorming} category clearly illustrates the limitations of linear modelling. The LME analysis yields minimal diagnostic insight, identifying only a single linear predictor: \textsc{Avg. Model Turn Duration} (Table~\ref{tab:combined_creativity_appendix}, Part 1). In contrast, the GAMM analysis uncovers significant non-linear predictors that characterize the creative interaction (Part 2). Satisfaction in this context is not driven by linear efficiency, but is instead associated with the user's iterative trajectory (\textsc{User Self-Consistency}) and the final state of the exploration (\textsc{Final Turn Distance from Goal}). These findings suggest that linear models are insufficient for open-ended, exploratory tasks, where satisfaction is governed by complex, user-centric interaction dynamics.

\begin{table}[h]
\centering\small
\caption{Linear and Non-Linear Effects for \textit{Creativity \& Brainstorming}}
\label{tab:combined_creativity_appendix}
\begin{threeparttable}
\begin{tabular}{p{6.5cm} S[table-format=1.4] l@{} S[table-format=1.4] c}
\toprule
\textbf{Feature} & \multicolumn{2}{c}{\textbf{Coefficient ($\beta$)}} & {\textbf{Std. Error}} & \textbf{\textit{p}-value} \\
\midrule
\multicolumn{5}{l}{\textit{\textbf{Part 1: Significant Linear Effects (LME)}}} \\
\addlinespace[0.3em]
\textsc{Avg. Model Turn Duration} & 0.0147 & * & 0.0062 & 0.017 \\
\midrule 
\multicolumn{5}{l}{\textit{\textbf{Part 2: Additional Significant Non-Linear Effects (GAMM)}}} \\
\addlinespace[0.3em]
\textsc{User Self-Consistency} & \multicolumn{2}{c}{---} & {---} & {$<0.001$} \\
\textsc{Final Turn Distance from Goal} & \multicolumn{2}{c}{---} & {---} & 0.008 \\
\textsc{User Adherence to Goal} & \multicolumn{2}{c}{---} & {---} & 0.011 \\
\bottomrule
\end{tabular}
\end{threeparttable}
\end{table}

\subsection{Troubleshooting \& Assistance}
In the context of \textit{Troubleshooting}, the linear model highlights the primacy of efficiency; satisfaction is negatively correlated with conversation length and repetition, while being positively associated with \textsc{Avg. Model Turn Duration}. The GAMM analysis (Table~\ref{tab:combined_troubleshooting_appendix}) reinforces these findings while elucidating the importance of the problem-solving process. Significant non-linear effects are observed for metrics quantifying the structural integrity of the dialogue, including \textsc{Conversation Drift from Goal}, \textsc{User Self-Consistency}, and \textsc{Initial Response Distance}. These results indicate that satisfaction in troubleshooting is a function not merely of the outcome, but of a structured, logical, and effectively initiated diagnostic trajectory.

\begin{table}[h]
\centering\small
\caption{Linear and Non-Linear Effects for \textit{Troubleshooting \& Assistance}}
\label{tab:combined_troubleshooting_appendix}
\begin{threeparttable}
\begin{tabular}{p{6.5cm} S[table-format=-1.4] l@{} S[table-format=1.4] c}
\toprule
\textbf{Feature} & \multicolumn{2}{c}{\textbf{Coefficient ($\beta$)}} & {\textbf{Std. Error}} & \textbf{\textit{p}-value} \\
\midrule
\multicolumn{5}{l}{\textit{\textbf{Part 1: Significant Linear Effects (LME)}}} \\
\addlinespace[0.3em]
\textsc{Max Model Self-Similarity}\tnote{\dag} & -6.3969 & *** & 1.2124 & {$<0.001$} \\
\textsc{Avg. Model Turn Duration}\tnote{\dag} & 0.0526 & *** & 0.0110 & {$<0.001$} \\
\textsc{Number of Turns}\tnote{\dag} & -0.1086 & *** & 0.0228 & {$<0.001$} \\
\textsc{Max Model Distance from User} & -3.4389 & *** & 1.0389 & {$<0.001$} \\
\textsc{Avg. Model Distance from User} & -4.3101 & ** & 1.4078 & 0.002 \\
\textsc{Trend in Goal Adherence}\tnote{\dag} & 17.7736 & ** & 6.8994 & 0.010 \\
\midrule 
\multicolumn{5}{l}{\textit{\textbf{Part 2: Additional Significant Non-Linear Effects (GAMM)}}} \\
\addlinespace[0.3em]
\textsc{Conversation Drift from Goal} & \multicolumn{2}{c}{---} & {---} & {$<0.001$} \\
\textsc{User Self-Consistency} & \multicolumn{2}{c}{---} & {---} & {$<0.001$} \\
\textsc{Median Gap Time} & \multicolumn{2}{c}{---} & {---} & {$<0.001$} \\
\textsc{Mad Gap Time} & \multicolumn{2}{c}{---} & {---} & {$<0.001$} \\
\textsc{Goal vs. Initial Prompt Distance} & \multicolumn{2}{c}{---} & {---} & {$<0.001$} \\
\textsc{Goal Convergence Ratio} & \multicolumn{2}{c}{---} & {---} & 0.007 \\
\textsc{Max User Distance from Model} & \multicolumn{2}{c}{---} & {---} & 0.014 \\
\textsc{Initial Response Distance} & \multicolumn{2}{c}{---} & {---} & 0.020 \\
\textsc{Model Self-Similarity} & \multicolumn{2}{c}{---} & {---} & 0.024 \\
\textsc{Avg. User Turn Duration} & \multicolumn{2}{c}{---} & {---} & 0.034 \\
\textsc{User Adherence to Goal} & \multicolumn{2}{c}{---} & {---} & 0.037 \\
\bottomrule
\end{tabular}
\begin{tablenotes}
    \item[\dag] \scriptsize\textit{Indicates a feature with a significant linear effect that also showed a significant non-linear effect.}
\end{tablenotes}
\end{threeparttable}
\end{table}

\subsection{Accessibility Support}
In the domain of \textit{Accessibility Support}, the linear model shows the importance of efficiency and focus; satisfaction correlates with direct, non-repetitive, and goal-oriented interactions (Table~\ref{tab:combined_accessibility_appendix}). The GAMM analysis refines this view by revealing a heightened sensitivity to conversational stability. The dominance of non-linear predictors such as \textsc{Trend in Model Relevance} and \textsc{Conversation Volatility} suggests that satisfaction in this context is fragile; it is heavily penalized by any evidence of degrading relevance or unpredictability. These findings identify consistency and sustained coherence as core prerequisites for user trust.

\begin{table}[h]
\centering\small
\caption{Linear and Non-Linear Effects for \textit{Accessibility Support}}
\label{tab:combined_accessibility_appendix}
\begin{threeparttable}
\begin{tabular}{p{6.5cm} S[table-format=-1.4] l@{} S[table-format=1.4] c}
\toprule
\textbf{Feature} & \multicolumn{2}{c}{\textbf{Coefficient ($\beta$)}} & {\textbf{Std. Error}} & \textbf{\textit{p}-value} \\
\midrule
\multicolumn{5}{l}{\textit{\textbf{Part 1: Significant Linear Effects (LME)}}} \\
\addlinespace[0.3em]
\textsc{Max Model Self-Similarity}\tnote{\dag} & -7.0134 & *** & 1.0150 & {$<0.001$} \\
\textsc{Model Self-Similarity}\tnote{\dag} & -6.1539 & *** & 1.1987 & {$<0.001$} \\
\textsc{Model Adherence to Goal} & 5.7163 & *** & 1.6100 & {$<0.001$} \\
\textsc{Model Adherence to Initial Prompt} & 4.0322 & *** & 1.1850 & {$<0.001$} \\
\textsc{Number of Turns}\tnote{\dag} & -0.0657 & ** & 0.0240 & 0.006 \\
\textsc{Final Model Response to Goal Distance} & 2.8064 & * & 1.1269 & 0.013 \\
\textsc{Max Model Distance from Goal} & 3.5097 & * & 1.4549 & 0.016 \\
\textsc{Avg. User Distance from Model}\tnote{\dag} & 3.5937 & * & 1.5685 & 0.022 \\
\textsc{Goal vs. Initial Prompt Distance} & 1.9109 & * & 0.8784 & 0.030 \\
\textsc{Min Model Distance to Goal} & 3.1272 & * & 1.4377 & 0.030 \\
\textsc{Avg. Model Distance from User}\tnote{\dag} & -3.1975 & * & 1.5038 & 0.033 \\
\midrule 
\multicolumn{5}{l}{\textit{\textbf{Part 2: Additional Significant Non-Linear Effects (GAMM)}}} \\
\addlinespace[0.3em]
\textsc{Trend in Model Relevance} & \multicolumn{2}{c}{---} & {---} & {$<0.001$} \\
\textsc{Conversation Volatility} & \multicolumn{2}{c}{---} & {---} & {$<0.001$} \\
\textsc{Avg. User Turn Duration} & \multicolumn{2}{c}{---} & {---} & {$<0.001$} \\
\textsc{Median Gap Time} & \multicolumn{2}{c}{---} & {---} & {$<0.001$} \\
\textsc{Mad Gap Time} & \multicolumn{2}{c}{---} & {---} & {$<0.001$} \\
\textsc{Min Model Distance to User Prompt} & \multicolumn{2}{c}{---} & {---} & {$<0.001$} \\
\textsc{Late Conversation Volatility} & \multicolumn{2}{c}{---} & {---} & 0.048 \\
\bottomrule
\end{tabular}
\begin{tablenotes}
    \item[\dag] \scriptsize\textit{Indicates a feature with a significant linear effect that also showed a significant non-linear effect.}
\end{tablenotes}
\end{threeparttable}
\end{table}

\subsection{Casual Interaction \& Entertainment}
In the domain of \textit{Casual Interaction}, the linear model presents a complex profile, identifying positive coefficients for both divergence (\textsc{Avg. User Distance from Model}) and goal adherence. The GAMM analysis (Table~\ref{tab:combined_casual_appendix}) resolves this tension by highlighting the non-linear nature of engagement in this context. The dominance of predictors such as \textsc{Min Model Distance to User Prompt} suggests that satisfaction is not driven by a constant average of relevance, but rather by specific instances of high alignment. This indicates that casual interactions are characterized by `peak moments' of connection; a single highly relevant or resonant response can have a disproportionate positive impact on satisfaction, even within a trajectory that is otherwise explorative or meandering.

\begin{table}[h]
\centering\small
\caption{Linear and Non-Linear Effects for \textit{Casual Interaction}}
\label{tab:combined_casual_appendix}
\begin{threeparttable}
\begin{tabular}{p{6.5cm} S[table-format=-1.4] l@{} S[table-format=1.4] c}
\toprule
\textbf{Feature} & \multicolumn{2}{c}{\textbf{Coefficient ($\beta$)}} & {\textbf{Std. Error}} & \textbf{\textit{p}-value} \\
\midrule
\multicolumn{5}{l}{\textit{\textbf{Part 1: Significant Linear Effects (LME)}}} \\
\addlinespace[0.3em]
\textsc{Avg. User Distance from Model}\tnote{\dag} & 8.4028 & *** & 1.6175 & {$<0.001$} \\
\textsc{Max Model Self-Similarity}\tnote{\dag} & -6.1617 & *** & 1.2534 & {$<0.001$} \\
\textsc{Model Self-Similarity} & -5.6712 & *** & 1.3523 & {$<0.001$} \\
\textsc{Model Adherence to Goal}\tnote{\dag} & 6.0265 & *** & 1.5622 & {$<0.001$} \\
\textsc{Late Conversation Volatility} & 4.1458 & *** & 1.1149 & {$<0.001$} \\
\textsc{Conversation Volatility}\tnote{\dag} & 7.3616 & *** & 1.9898 & {$<0.001$} \\
\textsc{Semantic Cohesion}\tnote{\dag} & -7.3257 & *** & 2.0573 & {$<0.001$} \\
\textsc{Final Model Response to Goal Distance}\tnote{\dag} & 3.9207 & ** & 1.2386 & 0.002 \\
\textsc{Number of Turns}\tnote{\dag} & -0.0868 & ** & 0.0278 & 0.002 \\
\textsc{Max Model Distance from Goal} & 4.3363 & ** & 1.3941 & 0.002 \\
\textsc{Min Model Distance to Goal}\tnote{\dag} & 4.1567 & ** & 1.4319 & 0.004 \\
\textsc{Avg. Model Turn Duration}\tnote{\dag} & 0.0313 & * & 0.0127 & 0.014 \\
\textsc{Model Adherence to Initial Prompt} & 2.9870 & * & 1.3195 & 0.024 \\
\textsc{User Adherence to Goal} & 4.4233 & * & 2.2398 & 0.048 \\
\midrule 
\multicolumn{5}{l}{\textit{\textbf{Part 2: Additional Significant Non-Linear Effects (GAMM)}}} \\
\addlinespace[0.3em]
\textsc{Min Model Distance to User Prompt} & \multicolumn{2}{c}{---} & {---} & {$<0.001$} \\
\textsc{Goal vs. Initial Prompt Distance} & \multicolumn{2}{c}{---} & {---} & {$<0.001$} \\
\textsc{Final Turn Distance from Goal} & \multicolumn{2}{c}{---} & {---} & 0.018 \\
\bottomrule
\end{tabular}
\begin{tablenotes}
    \item[\dag] \scriptsize\textit{Indicates a feature with a significant linear effect that also showed a significant non-linear effect.}
\end{tablenotes}
\end{threeparttable}
\end{table}
\newpage
\subsection{Information Seeking \& Identification}
For the \textit{Information Seeking} category, the linear model associates satisfaction with direct, goal-oriented responses. The GAMM analysis (Table~\ref{tab:combined_infoseek_appendix}) broadens this perspective by identifying significant non-linear effects related to temporal dynamics. The statistical significance of \textsc{Median Gap Time} and \textsc{Mad Gap Time} suggests that the underlying rhythm and pacing of the dialogue exert a non-linear influence on user perception. These results imply that for straightforward information retrieval, satisfaction is shaped not only by semantic relevance but also by the temporal consistency of the interaction.

\begin{table}[H]
\centering\small
\caption{Linear and Non-Linear Effects for \textit{Information Seeking}}
\label{tab:combined_infoseek_appendix}
\begin{threeparttable}
\begin{tabular}{p{6.5cm} S[table-format=-1.4] l@{} S[table-format=1.4] c}
\toprule
\textbf{Feature} & \multicolumn{2}{c}{\textbf{Coefficient ($\beta$)}} & {\textbf{Std. Error}} & \textbf{\textit{p}-value} \\
\midrule
\multicolumn{5}{l}{\textit{\textbf{Part 1: Significant Linear Effects (LME)}}} \\
\addlinespace[0.3em]
\textsc{Max Model Self-Similarity} & -7.9087 & *** & 1.7427 & {$<0.001$} \\
\textsc{Min Model Distance to Goal} & 6.0156 & *** & 1.6713 & {$<0.001$} \\
\textsc{Number of Turns}\tnote{\dag} & -0.1217 & *** & 0.0352 & {$<0.001$} \\
\textsc{Model Adherence to Goal}\tnote{\dag} & 6.5272 & ** & 2.2560 & 0.004 \\
\textsc{Avg. User Distance from Model}\tnote{\dag} & 7.9406 & ** & 2.9165 & 0.006 \\
\textsc{Final Model Response to Goal Distance} & 3.8248 & ** & 1.4130 & 0.007 \\
\textsc{Trend in Goal Adherence} & 21.5767 & * & 9.0300 & 0.017 \\
\textsc{User Self-Consistency}\tnote{\dag} & 5.4412 & * & 2.2935 & 0.018 \\
\textsc{Avg. Model Turn Duration}\tnote{\dag} & 0.0482 & * & 0.0222 & 0.030 \\
\textsc{Conversation Drift from Goal} & 6.5770 & * & 3.2173 & 0.041 \\
\textsc{Goal Convergence Ratio} & -0.4305 & * & 0.2111 & 0.041 \\
\midrule 
\multicolumn{5}{l}{\textit{\textbf{Part 2: Additional Significant Non-Linear Effects (GAMM)}}} \\
\addlinespace[0.3em]
\textsc{Median Gap Time} & \multicolumn{2}{c}{---} & {---} & {$<0.001$} \\
\textsc{Mad Gap Time} & \multicolumn{2}{c}{---} & {---} & {$<0.001$} \\
\textsc{Avg. Model Distance from User} & \multicolumn{2}{c}{---} & {---} & {$<0.001$} \\
\textsc{Model Self-Similarity} & \multicolumn{2}{c}{---} & {---} & 0.047 \\
\bottomrule
\end{tabular}
\begin{tablenotes}
    \item[\dag] \scriptsize\textit{Indicates a feature with a significant linear effect that also showed a significant non-linear effect.}
\end{tablenotes}
\end{threeparttable}
\end{table}

\newpage
\section{Analysis of Satisfaction Ratings by Task Category}
\label{app:satisfaction_analysis}
This appendix presents an analysis of user satisfaction ratings stratified by use case category, conducted to investigate the potential influence of task type as a confounding variable. Table~\ref{tab:satisfaction_by_category} details the sample size, mean satisfaction, and standard deviation for the six most prevalent categories. The results demonstrate that mean satisfaction scores are comparable across categories, showing a lack of pronounced domain-specific bias. This consistency supports the validity of our reward model, suggesting that its predictive performance is not attributable to a simple topic-based heuristic, as the task category itself offers limited predictive signal for user satisfaction.

\begin{table}[H]
\centering
\caption{Summary of User Satisfaction Scores by Task Category}
\label{tab:satisfaction_by_category}
\begin{tabular}{l c c c}
\toprule
\textbf{Use Case Category} & \textbf{Count} & \textbf{Mean Satisfaction} & \textbf{Std. Deviation} \\
\midrule
Creativity \& Brainstorming & 176 & 2.77 & 1.08 \\
Accessibility Support & 417 & 2.55 & 1.18 \\
Information Seeking \& Identification & 142 & 2.54 & 1.21 \\
Learning \& Education & 646 & 2.47 & 1.16 \\
Casual Interaction \& Entertainment & 326 & 2.45 & 1.22 \\
Troubleshooting \& Assistance & 411 & 2.42 & 1.12 \\
\bottomrule
\end{tabular}
\end{table}

\clearpage
\section{LLM Baseline Details}
\label{app:llm_baseline}

To establish a strong baseline for comparison, we optimized our LLM model through iterative prompt engineering. We evaluated multiple configurations to find the most effective prompt for predicting user satisfaction from raw comparison transcripts.

Interestingly, we found that explicitly providing the user's stated goal (Template 2) actually led to a slight \textit{decrease} in predictive performance compared to a standard evaluation of the transcript alone (Template 1). We hypothesise that providing the explicit goal might cause the LLM to over-index on semantic topical relevance rather than the nuanced implicit signals of frustration or effort that are better captured by reading the interaction naturally.

Consequently, \textbf{Template 1} was selected as the final baseline used in the main results (Table~\ref{tab:pairwise_accuracy}).

It is important to note that this baseline employs a ``one-shot'' approach, assessing the entire transcript in a single inference call. While performance might be further improved by computationally intensive turn-by-turn analysis or chain-of-thought reasoning, we specifically chose this single-pass approach to mirror standard, scalable auto-raters currently used in large-scale analyses, such as those employed in ``How People Use ChatGPT'' \citep{chatterji2025people}.

\subsection{Final Baseline Prompt (Template 1)}
\begin{tcolorbox}[colback=white, colframe=gray!50!black, title=\textbf{Standard Evaluation Prompt}]
\small
\begin{verbatim}
You are an impartial and objective evaluator of user satisfaction in 
conversational AI interactions.
Your task is to carefully review the following transcript of an interaction 
between a User and an AI Model.
Based ONLY on this transcript, provide a numerical rating representing the 
User's likely satisfaction with the interaction.

Use the following 1-5 scale:
1: Very dissatisfied
2: Somewhat dissatisfied
3: Neither satisfied nor dissatisfied
4: Somewhat satisfied
5: Very satisfied

When making your assessment, consider factors such as:
- Was the AI Model's response relevant to the User's prompts?
- Did the AI Model seem to understand the User's intent?
- Was the User likely to achieve their goal by the end of the interaction?
- Was the interaction efficient, or did the User have to struggle?
- Was the AI Model polite and the tone appropriate?

Your response MUST be a single integer from 1 to 5 only. Do not provide 
any explanation, additional text, or punctuation.

TRANSCRIPT:
{transcript_text}

SATISFACTION SCORE (1-5):
\end{verbatim}
\end{tcolorbox}

\subsection{Alternative Goal-Aware Prompt (Template 2)}
\textit{Note: This prompt yielded lower performance during development and was not used for the final results.}
\begin{tcolorbox}[colback=white, colframe=gray!50!black, title=\textbf{Goal-Aware Evaluation Prompt}]
\small
\begin{verbatim}
You are an impartial and objective evaluator of user satisfaction in 
conversational AI interactions.
Your task is to carefully review the following transcript of an interaction 
between a User and an AI Model, along with the User's stated goal.
Based on this information, provide a numerical rating representing the 
User's likely satisfaction with the interaction.

When making your assessment, pay special attention to whether the AI model 
helped the user accomplish their stated goal.

USER'S GOAL:
{goal_text}

Use the following 1-5 scale:
1: Very dissatisfied
2: Somewhat dissatisfied
3: Neither satisfied nor dissatisfied
4: Somewhat satisfied
5: Very satisfied

Your response MUST be a single integer from 1 to 5 only. Do not provide 
any explanation, additional text, or punctuation.

TRANSCRIPT:
{transcript_text}

SATISFACTION SCORE (1-5):
\end{verbatim}
\end{tcolorbox}

\bibliography{main}

@techreport{chatterji2025people,
  title={How people use chatgpt},
  author={Chatterji, Aaron and Cunningham, Thomas and Deming, David J and Hitzig, Zoe and Ong, Christopher and Shan, Carl Yan and Wadman, Kevin},
  year={2025},
  institution={National Bureau of Economic Research}
}

@misc{gooding2025writingtestbedopenended,
      title={Writing as a testbed for open ended agents}, 
      author={Sian Gooding and Lucia Lopez-Rivilla and Edward Grefenstette},
      year={2025},
      eprint={2503.19711},
      archivePrefix={arXiv},
      primaryClass={cs.CL},
      url={https://arxiv.org/abs/2503.19711}, 
}

@misc{vajjala2025opportunitieschallengesllmseducation,
      title={Opportunities and Challenges of LLMs in Education: An NLP Perspective}, 
      author={Sowmya Vajjala and Bashar Alhafni and Stefano Bannò and Kaushal Kumar Maurya and Ekaterina Kochmar},
      year={2025},
      eprint={2507.22753},
      archivePrefix={arXiv},
      primaryClass={cs.CL},
      url={https://arxiv.org/abs/2507.22753}, 
}

@inproceedings{dong2023towards,
  title={Towards next-generation intelligent assistants leveraging llm techniques},
  author={Dong, Xin Luna and Moon, Seungwhan and Xu, Yifan Ethan and Malik, Kshitiz and Yu, Zhou},
  booktitle={Proceedings of the 29th ACM SIGKDD Conference on Knowledge Discovery and Data Mining},
  pages={5792--5793},
  year={2023}
}

@misc{fragiadakis2025evaluatinghumanaicollaborationreview,
      title={Evaluating Human-AI Collaboration: A Review and Methodological Framework}, 
      author={George Fragiadakis and Christos Diou and George Kousiouris and Mara Nikolaidou},
      year={2025},
      eprint={2407.19098},
      archivePrefix={arXiv},
      primaryClass={cs.HC},
      url={https://arxiv.org/abs/2407.19098}, 
}

@inproceedings{10.1145/3491102.3502030,
author = {Lee, Mina and Liang, Percy and Yang, Qian},
title = {CoAuthor: Designing a Human-AI Collaborative Writing Dataset for Exploring Language Model Capabilities},
year = {2022},
isbn = {9781450391573},
publisher = {Association for Computing Machinery},
address = {New York, NY, USA},
url = {https://doi.org/10.1145/3491102.3502030},
doi = {10.1145/3491102.3502030},
booktitle = {Proceedings of the 2022 CHI Conference on Human Factors in Computing Systems},
articleno = {388},
numpages = {19},
location = {New Orleans, LA, USA},
series = {CHI '22}
}

@inproceedings{mirowski2023co,
  title={Co-writing screenplays and theatre scripts with language models: Evaluation by industry professionals},
  author={Mirowski, Piotr and Mathewson, Kory W and Pittman, Jaylen and Evans, Richard},
  booktitle={Proceedings of the 2023 CHI conference on human factors in computing systems},
  pages={1--34},
  year={2023}
}

@misc{rafailov2024directpreferenceoptimizationlanguage,
      title={Direct Preference Optimization: Your Language Model is Secretly a Reward Model}, 
      author={Rafael Rafailov and Archit Sharma and Eric Mitchell and Stefano Ermon and Christopher D. Manning and Chelsea Finn},
      year={2024},
      eprint={2305.18290},
      archivePrefix={arXiv},
      primaryClass={cs.LG},
      url={https://arxiv.org/abs/2305.18290}, 
}
\end{document}